\documentclass{article}

\usepackage{microtype}
\usepackage{graphicx}
\usepackage{subfigure}
\usepackage{booktabs} 

\usepackage{hyperref}

\usepackage[accepted]{icml2023}

\usepackage{amsmath}
\usepackage{amssymb}
\usepackage{mathtools}
\usepackage{amsthm}
\usepackage{bm}
\usepackage{amsfonts}
\usepackage{enumerate}
\usepackage{multirow}
\usepackage{algorithmic}
\usepackage{arydshln}
\usepackage{subfigure}
\usepackage{enumitem}
\usepackage{physics}
\usepackage{algorithm}
\usepackage{threeparttable}
\usepackage{bbm}
\usepackage{booktabs}
\usepackage{epstopdf}
\usepackage{graphics}
\usepackage{epsfig}
\usepackage{url}

\usepackage[capitalize,noabbrev]{cleveref}

\theoremstyle{plain}
\newtheorem{theorem}{Theorem}[section]

\theoremstyle{definition}
\newtheorem{definition}[theorem]{Definition}

\theoremstyle{remark}

\newcommand{\clpl}[0]{\textsc{Clpl}}

\newcommand{\vpll}[0]{\textsc{Valen}}
\newcommand{\rcpll}[0]{\textsc{RC}}
\newcommand{\ccpll}[0]{\textsc{CC}}
\newcommand{\proden}[0]{\textsc{Proden}}
\newcommand{\cavl}[0]{\textsc{Cavl}}
\newcommand{\lws}[0]{\textsc{Lw}}

\newcommand{\fid}[0]{\textsc{Pop}}
\newcommand{\pico}[0]{\textsc{Pico}}
\newcommand{\rcr}[0]{\textsc{Rcr}}
\renewcommand{\algorithmicrequire}{\textbf{Input:}} 
\renewcommand{\algorithmicensure}{\textbf{Output:}}
\usepackage[textsize=tiny]{todonotes}

\icmltitlerunning{Progressive Purification for Instance-Dependent Partial Label Learning}

\begin{document}

\twocolumn[
\icmltitle{Progressive Purification for Instance-Dependent Partial Label Learning}



\icmlsetsymbol{equal}{*}

\begin{icmlauthorlist}
\icmlauthor{Ning Xu}{sch}
\icmlauthor{Biao Liu}{sch}
\icmlauthor{Jiaqi Lv}{comp}
\icmlauthor{Congyu Qiao}{sch}
\icmlauthor{Xin Geng}{sch}
\end{icmlauthorlist}

\icmlaffiliation{comp}{RIKEN Center for Advanced Intelligence Project, Tokyo 103-0027, Japan}
\icmlaffiliation{sch}{School of Computer Science and Engineering, Southeast University, Nanjing, China}

\icmlcorrespondingauthor{Ning Xu}{xning@seu.edu.cn}
\icmlcorrespondingauthor{Xin Geng}{xgeng@seu.edu.cn}

\icmlkeywords{Machine Learning, ICML}

\vskip 0.3in
]



\printAffiliationsAndNotice{} 

\begin{abstract}
Partial label learning (PLL) aims to train multi-class classifiers from the examples each annotated  with a set of candidate labels where a fixed but unknown candidate label is correct. In the last few years, the {instance-independent} generation process of candidate labels has been extensively studied, on the basis of which many  theoretical advances have been made in PLL. Nevertheless, the candidate labels are always \emph{instance-dependent}  in practice and  there is no theoretical guarantee  that the model trained on the instance-dependent PLL examples can converge to an ideal one.
In this paper, a theoretically grounded and practically effective  approach named  {\fid}, i.e. PrOgressive Purification for instance-dependent partial label learning, is proposed. Specifically,  {\fid} updates the learning model  and purifies each candidate label set progressively in every epoch.
Theoretically, we prove that {\fid} enlarges the region appropriately fast where the model is {reliable}, and eventually approximates the Bayes optimal classifier with mild assumptions. Technically, {\fid} is flexible with arbitrary PLL {losses} and could improve the performance of the previous  PLL losses in the instance-dependent case. Experiments on the benchmark datasets and the real-world datasets validate the effectiveness of the proposed method. 
\end{abstract}

\section{Introduction}\label{sec1}

The difficulty of collecting large scale datasets with high-quality annotations for training  classifiers  induces weakly supervised learning, a typical example among which is {partial label learning} (PLL) \cite{Nguyen_Caruana2008,Cour_2011,zhang2017disambiguation,yao2020network,lv2020progressive,feng2020provably,wen2021leveraged}. PLL deals with the problem where each training example is associated with
a set of candidate labels, among which only one label is valid. This paradigm naturally arises in
various real-world applications, such as web mining \cite{Jie_Orabona2010}, multimedia content analysis \cite{Zeng_2013}, ecoinformatics \cite{Tang_Zhang2017}, etc.

A large number of deep PLL algorithms have recently emerged that aimed to design regularizers \cite{yao2020deep,yao2020network,lyu2022partial} or network architectures \cite{wang2022pico} for PLL data.
Further, there are some PLL works that provided theoretical guarantees while making their methods compatible with deep networks \cite{lv2020progressive,feng2020provably,wen2021leveraged,wu2021learning}. 
These existing  works have focused on the \emph{instance-independent} setting where the generation process of candidate labels is homogeneous across training examples.
With an explicit formulation of the generation process, the {asymptotical consistency} \cite{mohri2018foundations} of the methods, i.e., the classifier learned from partially labeled examples could approximate the Bayes optimal classifier, can be analyzed.

 Previous works have extensively studied  instance-independent PLL and many  theoretical advances have been made in this setting. However, the candidate labels are always instance-dependent (feature-dependent) in practice as the incorrect labels related to the feature  are more likely to  be picked as candidate label set for each instance. Therefore,  \emph{instance-dependent} (ID) candidate labels \cite{xu2021instance} should be quite realistic and could describe the ambiguous labeling information for the  instance which is difficult to be annotated with an exact true label in PLL.  By adopting the latent label distributions, recent work \cite{xu2021instance}  has  empirically validated that the classifier trained on  instance-dependent PLL examples  could achieve good performance. Nevertheless, there is no theoretical guarantee  that the model trained on the instance-dependent PLL examples can converge to an ideal one.

 In this paper, we propose a theoretically grounded  method  named {\fid}, i.e. PrOgressive Purification for instance-dependent partial label learning. Specifically,  the observed candidate labels are utilized to train a randomly initialized classifier (deep network) for several epochs, and then the classifier is updated with purified candidate label set  for the remaining epochs. 
In each epoch,  each candidate label  set is purified according to the pure level set with the classifier for candidate labels and we  prove that {\fid} can be guaranteed to enlarge the region where the model is {reliable} by a promising rate. As a consequence, the false candidate labels are gradually moved out and the classification performance of the classifier is improved.  We justify {\fid} and outline the main contributions below:
\begin{itemize}[topsep=0ex,leftmargin=*,parsep=1pt,itemsep=1pt]	
	\item We propose a novel approach named {\fid} for the instance-dependent PLL problem, which purifies the candidate label sets and refines the classifier iteratively. Extensive experiments validate the effectiveness of {\fid}.
	\item We prove that {\fid} can be guaranteed to enlarge the region where the model is {reliable} by a promising rate, and eventually approximates the Bayes optimal classifier with mild assumptions.  To the best of our knowledge, this is the first theoretically guaranteed approach for instance-dependent PLL.
	\item {\fid} is flexible with respect to {losses}, so that the losses designed for the instance-independent PLL problems can be embedded directly. We empirically show that such embedding allows advanced PLL losses can be applied to the instance-dependent PLL problem and achieve state-of-the-art learning performance.
\end{itemize}

\section{Related Work}
In this section, we briefly go through the seminal works in PLL, focusing on the theoretical works and discussing the underlying assumptions behind them.

There have been substantial traditional PLL algorithms from the pioneering work~\cite{Jin_Ghahramani2003}.
From a practical standpoint, they have been studied along two different research routes: the identification-based strategy and the average-based strategy. 
The identification-based strategy purifies each partial label and extracts the true label heuristically in the training phase, so as to identify the true labels \cite{Chen_2014,Zhang_ZhouBin2016,Tang_Zhang2017,feng2019partial,xu2019partial}.
On the contrary, the average-based strategy treats all candidates equally \cite{Hullermeier_Beringer2006,Cour_2011,ZY15}.
On the theoretical side, Liu and Dietterich~\cite{liu2012conditional} analyzed the learnability of PLL by making a {small ambiguity degree condition} assumption, which ensures classification errors on any instance have a probability of being detected.
And Cour \textit{et al.}~\cite{Cour_2011} proposed a consistent approach under the small ambiguity degree condition and a dominance assumption on data distribution.
Liu and Dietterich~\cite{liu2012conditional} proposed a Logistic Stick-Breaking Conditional Multinomial Model to portray the mapping between instances and true labels while assuming the generation of the partial label is independent of the instance itself. The label distribution  is adopted to disambiguate the candidate labels \cite{xu2019partial} via recovering the latent label distribution in the label enhancement process \cite{xu2019label,xu2022one,XuVariational2023}.
It should be noted that the vast majority of traditional PLL works have only empirically verified the performance of algorithms on small data sets, without formalizing the statistical model for the PLL problem, and therefore even less so for theoretical analysis of when and why the algorithms work.

In recent years, deep learning  has been applied to PLL and has greatly advanced the practical application of PLL.
Yao \textit{et al.}~\cite{yao2020deep,yao2020network} and Lv \textit{et al.}~\cite{lv2020progressive} proposed learning objectives that are compatible with stochastic optimization and thus can be implemented by deep networks.
Soon Feng \textit{et al.} \cite{feng2020provably} formalized the first generation process for PLL.
They assumed that given the latent true label, the probability of all incorrect labels being added into the candidate label set is uniform and independent of the instance.
Thanks to the uniform generation process, they proposed two provably consistent algorithms.
Wen \textit{et al.} \cite{wen2021leveraged} extended the uniform one to the {class-dependent} case, but still keep the instance-independent assumption unchanged.
In addition, a new paradigm called complementary label learning \cite{ishida2017learning,yu2018learning,ishida2019complementary,feng2020learning} has been proposed that learns from instances equipped with a complementary label. 
A complementary label specifies the classes to which the instance does not belong, so it can be considered to be an inverted PLL problem.
However, all of them made the instance-independent assumption for analyzing the statistic consistency.
Wu and Sugiyama~\cite{wu2021learning} proposed a framework that unifies the formalization of multiple generation processes under the instance-independent assumption. Wang \textit{et al.} \cite{wang2022pico} proposed a data-augmentation-based framework to disambiguate partial labels with contrastive learning. Zhang \textit{et al.} \cite{zhang2021exploiting} exploited the class activation value to identify the true label in candidate label sets.

Very recently, some researchers are beginning to notice a more general setting--- instance-independent (ID) PLL.
Learning with the ID partial labels is challenging, and all instance-independent approaches cannot handle the ID PLL problem directly.
Specifically, the theoretical approaches mentioned above utilize mainly the {loss correction} technique, which corrects the prediction or the loss of the classifier using a {prior or estimated knowledge} of data generation processes, i.e., a set of parameters controlling the probability of generating incorrect candidate labels, or it is often called transition matrix \cite{patrini2017making}.
The transition matrix can be characterized fixedly in the instance-independent setting since it does not need to include instance-level information, a condition that does not hold in ID PLL.
Furthermore, it is {ill-posed} to estimate the transition matrix by only exploiting partially labeled data, i.e., the transition matrix is unidentifiable \cite{xia2020part}.
Therefore, some new methods should be proposed to tackle this issue.
Xu \textit{et al.}~\cite{xu2021instance} introduced a solution that infers the latent label posterior via variational inference methods \cite{blei2017variational}, nevertheless, its effectiveness would be hardly guaranteed.
In this paper, we propose {\fid} for the ID PLL problem and theoretically prove that the learned classifier approximates well to the Bayes optimal.

\section{Proposed Method}
\subsection{Preliminaries}\label{preliminaries}
First of all, we briefly introduce some necessary notations. 
Consider a multi-class classification problem of $c$ classes. 
Let $\mathcal{X}=\mathbb{R}^q$ be the $q$-dimensional instance space and $\mathcal{Y}=\{1,2,\ldots,c\}$ be the label space with $c$ class labels. 
In supervised learning, let $p(\boldsymbol{x},y)$ be the underlying ``clean'' distribution generating $(\boldsymbol{x},y^{\boldsymbol{x}})\in\mathcal{X}\times\mathcal{Y}$ from which $n$ i.i.d.~samples $\{(\boldsymbol{x}_i,y^{\boldsymbol{x}_i})\}^n_{i=1}$ are drawn. 

In PLL, there is a {candidate label space} $\mathcal{S}:=\{S|S\subseteq\mathcal{Y},S\neq\emptyset\}$ and the PLL training set $\mathcal{D}=\{(\bm{x}_i,S_i)|1\leq i\leq n\}$ is sampled independently and identically from a ``corrupted'' density $\tilde{p}(\boldsymbol{x},S)$ over $\mathcal{X}\times\mathcal{S}$.
It is generally assumed that $p(\boldsymbol{x},y)$ and $p(\boldsymbol{x},S)$ have the same marginal distribution of instances $p(\boldsymbol{x})$.
Then the {generation process} of partial labels can thus be formalized as $p(S|\boldsymbol{x})=\sum_{y}p(S|\boldsymbol{x},y)p(y|\boldsymbol{x})$.
We define the probability that, given the instance $\bm{x}$ and its class label $y^{\bm{x}}$, $j$-label being included in its partial label as the {flipping probability}:
\begin{equation*}
	\xi^j(\bm{x})=p(j\in S|\bm{x},y^{\bm{x}}), \ \forall j\in\mathcal{Y},
\end{equation*} 
The key definition in PLL is that the latent true label of an instance is always one of its candidate label, i.e., $\xi^{y^{\bm{x}}}(\bm{x})=1$.

We consider use deep models by the aid of an inverse link function \cite{reid2010composite} $\boldsymbol{\phi}:\mathbb{R}^c\rightarrow\Delta^{c-1}$ where $\Delta^{c-1}$ denotes the $c$-dimensional simplex, for example, the softmax, as learning model in this paper.
Then the goal of supervised multi-class classification and PLL is the same: a scoring function $f:\mathcal{X}\mapsto\Delta^{c-1}$ that can make correct predictions on unseen inputs.
Typically, the classifier takes the form:
\begin{equation*}
	h(\boldsymbol{x})=\arg\max_{j\in \mathcal{Y}}f_j(\boldsymbol{x}).
\end{equation*}
The Bayes optimal classifier $h^\star$ (learned using supervised data) is the one that minimizes the risk w.r.t the 0-1 loss (or some classification-calibrated loss \cite{bartlett2006convexity}), i.e.,
\begin{equation*}
	h^\star=\arg\min_{h}\mathcal{R}_{01}=\arg\min_{h}\mathbb{E}_{(\boldsymbol{X},Y)\sim p(\boldsymbol{x},y)}\big[\mathbf{1}_{\{h(\boldsymbol{X})\neq Y\}}\big].
\end{equation*} 
For \emph{strictly proper losses} \cite{gneiting2007strictly}, the scoring function $f^*$ recovers the class-posterior probabilities, i.e., $f^\star(\boldsymbol{x})=p(y|\boldsymbol{x}), \forall \boldsymbol{x}\in\mathcal{X}$.
When the supervision information available is partial label, the PLL risk under $\tilde{p}(\boldsymbol{x},S)$ w.r.t.~a suitable {PLL loss} $\mathcal{L}:\mathbb{R}^k\times\mathcal{S}\rightarrow\mathbb{R}^+$ is defined as
\begin{equation*}
	\tilde{\mathcal{R}}=\mathbb{E}_{(\boldsymbol{X},S)\sim\tilde{p}(\boldsymbol{x},S)}\big[\mathcal{L}(h(\boldsymbol{X}),S)\big].
\end{equation*}
Minimizing $\tilde{\mathcal{R}}$ induces the classifier and it is desirable that the minimizer approach $h^\star$.
In addition, let $o=\arg\max_{j\neq y^{\bm{x}}} p(y=j|\bm{x})$ be the class label with the second highest posterior possibility among all labels. 

\subsection{Overview}
In the latter part of this section, we will introduce a concept \emph{pure level set} as the region where the model is {reliable}. 
We prove that given a tiny reliable region, one could progressively enlarge this region and improves the model with a sufficient rate by disambiguating the partial labels.
Motivated by the theoretical results, we propose an approach {\fid} that works by progressively purifying the partial labels to move out the false candidate labels, and eventually the learned classifier could approximate the Bayes optimal classifier.

{\fid} employs the observed partial labels to pre-train a randomly initialized classifier for several epochs, and then  updates both partial labels and the classifier for the remaining epochs.  
We start with a warm-up period, in which we train the predictive model with a well-defined PLL loss \cite{lv2020progressive}. 
This allows us to attain a reasonable predictive model before it starts fitting incorrect labels \cite{ZhangBHRV17}. 
After the warm-up period, we iteratively purify each partial label by moving out the candidate labels for which the current classifier has high confidence of being incorrect, and subsequently we train the classifier with the purified partial labels in the next epoch.  
After the model has been fully trained, the predictive model can perform prediction for unseen instances.

\subsection{The {\fid} Approach}\label{Theoretical}

We assume that the hypothesis class $\mathcal{H}$ is sufficiently complex (and deep networks could meet this condition), such that the approximation error equals zero, i.e., $\arg\min_{h}\mathcal{R}=\arg\min_{h\in\mathcal{H}}\mathcal{R}$ and we have enough training data i.e., $ n \rightarrow \infty $. The classifier is able to at least approximate the Bayes optimal classifier $ h^\star $ and the gap between the learned  $f(\bm{x})$ and the the scoring function $f^\star(\bm x)$ corresponding to  $ h^\star $  is determined by the inconsistency between incorrect candidate labels and output of the Bayes optimal classifier. 

For two instance $\bm{x}$ and $\bm{z}$ that satisfy $p(y^{\bm{z}}|\bm{z})- p(o|\bm{z})\geq p(y^{\bm{x}}|\bm{x})- p(o|\bm{x})$, i.e.,  the margin between the posterior of ground-truth label $ p(y^{\bm z} \vert \bm z) $  and the second highest posterior possibility $ p(o \vert \bm z) $  is larger than that in point $ \bm x $, the indicator function $\big[\mathbf{1}_{\{{ {j\neq h^\star(\bm z)}}\}}\Big\vert {}p(y^{\bm{z}}|\bm{z})- p(o|\bm{z})\geq p(y^{\bm{x}}|\bm{x})- p(o|\bm{x}), ~j \in S_{\bm z}\big]$ equals 1 if  the candidate label $j$  of $ \bm z $  is inconsistent with the output of the optimal Bayes classifier $h^\star(\bm z)$. 
Then, the gap between  $f_j(\bm{x})$ and {$ f^\star_j(\bm x) $ }, i.e.,  the approximation error of the classifier, could be controlled by the inconsistency between the incorrect candidate labels and the output of  the Bayes optimal classifier $ h^\star $ for all the instances  $\bm{z}$. Therefore, we assume that there exist constants $\alpha$, $\epsilon < 1$, such that for $f(\bm{x})$,
\begin{equation}\label{gap}
	\begin{split}
		\vert f^j(\bm{x})-p(y=j|\bm{x})\vert\leq  \alpha\mathbb{E}_{(\bm{z},S)\sim \tilde{p}(\bm{z},S)} \bigg[\mathbf{1}_{\{j\neq y^{\bm{z}} \}} \Big\vert \\ p(y^{\bm{z}}|\bm{z}) - p(o|\bm{z})\geq p(y^{\bm{x}}|\bm{x})- p(o|\bm{x})\bigg]+\frac{\epsilon}{6}
	\end{split}
\end{equation}
where  the scoring function $f^*$ corresponding to  $h^*$ on \emph{strictly proper losses} \cite{gneiting2007strictly} recovers the class-posterior probabilities, i.e., $f^\star_j(\boldsymbol{x})=p(y=j|\boldsymbol{x})$. In addition, for the probability density function $d(u)$ of cumulative distribution function $ D(u)=P_{\bm x\sim p(\bm x,y)}(u(\bm x) \leq u) $ where $ 0 \leq u \leq 1 $ and the margin $u(\bm{x})= p(y^{\bm{x}}|\bm{x})- p(o|\bm{x}) $. we assume that there exist constants $c_{\star}$, $c^{\star} > 0$ such that $c_{\star} < d(u) < c^{\star}$. Then,  the worst-case density-imbalance ratio is denoted by $l=\frac{c^\star}{c_\star}$. As the flipping probability of the incorrect label  in the instance-dependent generation process  is related to its posterior probability, we assume that there exists a constant $t>0$ such that:
\begin{equation}
	\xi^j(\bm{x}) \leq  p(y=j|\bm{x})t.
\end{equation}

Motivated by the pure level set in binary classification \cite{ZhangZW0021}, we   define the pure level set in instance-dependent PLL, i.e.,  the region where the model is reliable:
\begin{definition}
	(Pure $(e, f)$-level set). A set $L(e):=\left\{\boldsymbol{x} \|  p(y^{\bm{x}}|\bm{x})- p(o|\bm{x}) \mid \geq e\right\}$ is pure for $f$ if $y^{\bm{x}}=\arg\max_{j} f_j(\bm x)$ for all $\boldsymbol{x} \in L(e)$.
\end{definition}
Assume that  there exists a set $ L(e) $ for all $\bm{x} \in L(e)$ which satisfies $y^{\bm{x}}=\arg\max_{j} f_j(\bm x)$, we have
\begin{equation}
	\begin{split}
		\mathbb{E}_{(\bm{z},S)\sim \tilde{p}(\bm{z},S)}\bigg[\mathbf{1}_{\{ {j \neq h^\star(\bm z)} \}}\Big\vert  p(y^{\bm{z}}|\bm{z}) \\ - p(o|\bm{z})\geq p(y^{\bm{x}}|\bm{x})- p(o|\bm{x}) {,~j \in S_{\bm z}} \bigg]=0	
	\end{split}
\end{equation}

which means that there is a tiny region $L(e):=\left\{\boldsymbol{x} \|  p(y^{\bm{x}}|\bm{x})- p(o|\bm{x}) \mid \geq e\right\}$ where  the model $f$ is reliable.

Let  $e_\text{new}$ be the new boundary and $\frac{\epsilon}{6l\alpha}( p(y^{\bm{x}}|\bm{x})-e)\leq e-e_\text{new}\leq\frac{\epsilon}{3l\alpha}( p(y^{\bm{x}}|\bm{x})-e)$. As the probability density function $d(u)$ of the margin $u(\bm{x})= p(y^{\bm{x}}|\bm{x})- p(o|\bm{x}) $ is bounded by  $c_{\star} < d(u) < c^{\star}$, we have the following result for $\bm{x}$ that satisfies $ e > p(y^{\bm{x}}|\bm{x})- p(o|\bm{x}) \geq e_\text{new}$ \footnote{More details could be found in Appendix A.1.}:
\begin{equation}\label{e_new}
	\begin{aligned}
		\mathbb{E}_{(\bm{z},S)\sim \tilde{p}(\bm{z},S)}\bigg[\mathbf{1}_{\{ {j \neq h^\star(\bm z)} \}}\Big\vert  p(y^{\bm{z}}|\bm{z})- \\ p(o|\bm{z})\geq p(y^{\bm{x}}|\bm{x})- p(o|\bm{x}) {,~j \in S_{\bm z}} \bigg] \leq \frac{\epsilon}{3\alpha}.
	\end{aligned}
\end{equation}

Combining Eq. (\ref{gap}) and Eq. (\ref{e_new}), there is 
\begin{equation}\label{}
	\begin{aligned}
		\vert f_j(\bm{x})-{f^\star_j(\bm x)}\vert \leq \frac{\epsilon}{2}.
	\end{aligned}
\end{equation}
Denote by $m=\arg\max_j f_j(\bm{x})$ the label with the highest posterior probability for the current prediction.
If $f_m(\bm{x})-f_{j\neq m}(\bm{x})\geq e+\epsilon$, we have \footnote{More details could be found in Appendix A.2.}
\begin{equation}\label{}
	\begin{aligned}
		p(y^{\bm{x}}|\bm{x}) \geq  p(y=j|\bm{x}) + e
	\end{aligned}
\end{equation}
which means that the label $j$ is incorrect label. Therefore, we could move the label $j$ out from the candidate label set to disambiguate the partial label, and then refine the learning model with the partial label with less ambiguity. In this way, we would move one step forward by trusting the model with the tiny reliable region with following theorem.

We start with a warm-up period, as the classifier is able to attain reasonable outputs before fitting label noise \cite{ZhangBHRV17}. Note that the warm-up training is employed to find a tiny reliable region  and the ablation experiments show that the performance of {\fid} does not rely on the warm-up strategy. The predictive model $\bm{\theta}$ could be trained on partially labeled examples   by minimizing  any   PLL  loss function. Here we adopt {\proden} loss \cite{lv2020progressive} to to find a tiny reliable region:
\begin{equation}\label{minimal}
	\mathcal{L}_{PLL}=\sum_{i=1}^{n}\sum_{j=1}^{c}w_{ij}\ell(f_j(\bm{x}_i), S_i).
\end{equation}
Here, $\ell$ is the cross-entropy loss and the weight  $w_{ij}$  is initialized with with uniform weights
and then  could be tackled  simply using the current predictions for
slightly putting more weights on more possible labels  \cite{lv2020progressive}:
\begin{equation}
	w_{ij}=\left\{\begin{array}{cl}
		f_{j}\left(\boldsymbol{x}_{i} \right) / \sum_{j \in S_{i}} f_{j}\left(\boldsymbol{x}_{i} \right) & \text { if } {j} \in S_{i} \\
		0 & \text { otherwise }
	\end{array}\right.
\end{equation}

\begin{algorithm}[t]
	\caption{{\fid} Algorithm}\label{algorithm}
	\textbf{Input}: The PLL training set $\mathcal{D}=\{(\bm{x}_1,S _1),...,(\bm{x}_n,S_n)\}$, initial threshold  $e_0$,  end threshold $e_\text{end}$, total round $R$, step-size $e_s$;
	\begin{algorithmic}[1]
		\STATE Initialize the predictive model $\bm{\theta}$ by warm-up training with the PLL loss Eq. \ref{minimal}, and threshold $e=e_0$;
		\FOR{$r=1,...,R$}
		\STATE Train the predictive model $f$ on $\mathcal{D}$;
		\FOR{$i=1,...,n$} 
		\FOR{$j\in S_i$}
		\IF{$f_{m_i}(\bm{x}_i)-f_j(\bm{x}_i) \geq e+\epsilon$}
		\STATE Purify the incorrect label $j$ by removing it from the candidate label set $S_i$;
		\ENDIF
		\ENDFOR
		\ENDFOR
		\IF{$e \leq e_\text{end}$, and there is no purification for any candidate label set}
		\STATE Decrease $e$ with step-size $e_s$;
		\ENDIF
		\ENDFOR
	\end{algorithmic}
	\textbf{Output}:  The final predictive model $f$ 
\end{algorithm}

\begin{theorem}\label{theorem1}
	Assume that {we have enough training data($n\rightarrow\infty$) and } there is a pure $ (e,f) $-level set where $ \bm x \in L(e) $ can be correctly classified by $ f $. For each $ \bm{x} $ and  $\forall j \in S$ and $j \neq m$,  if $ f_m(\bm{x})-f_j(\bm{x})\geq e+\epsilon$, we move out label $j$ from the candidate label set and then update the candidate label set as $S_\text{new}$. Then the new classifier $f_{\text{new}}(\bm{x})$ is trained on the updated data with  the new  distribution $ \tilde{p}(\bm{x},S_\text{new})$. Let $e_{\text{new}}$ be the minimum boundary that $L(e_\text{new})$ is pure for $f_\text{new}$. Then, we have
	
	$$p(y^{\bm{x}}|\bm{x})-e_{\text{new}}\geq(1+\frac{\epsilon}{6\alpha l})(p(y^{\bm{x}}|\bm{x})-e).$$
\end{theorem}
The detailed proof can be found in Appendix A.1.
Theorem \ref{theorem1} shows that the purified region $\gamma = p(y^{\bm{x}}|\bm{x})-e$ would be enlarged by  at least a constant factor with the given purification strategy. 

After the warm-up period, the classifier could be employed for  purification. According to Theorem \ref{theorem1}, we could progressively move out the incorrect candidate label with the continuously strict bound, and subsequently  train an effective  classifier with the purified  labels with the PLL loss \cite{lv2020progressive} since the PLL loss \cite{lv2020progressive} is model-independent and could operates in a mini-batched training manner to update  the model with the labeling-confidence weight. Specifically, we set a high threshold $e_0$ and calculate the difference $f_{m}(\bm{x}_i)-f_j(\bm{x}_i)$ for each candidate label. If there is a label $j$ for $\bm{x}_i$ satisfies $f_{m}(\bm{x}_i)-f_j(\bm{x}_i)\geq e_0$, we  move out it from the candidate label set and update the candidate label set. We depart from the theory by reusing the same fixed dataset over and over, but the empirics are reasonable.

If there is no purification for all partial labels, we begin to decrease the threshold $e$ and continue the purification for improving the training of the model. In this way, the incorrect candidate labels are progressively removed from the partial label round by round, and the performance of the classifier is continuously improved.  The algorithmic description of {\fid} is shown in Algorithm \ref{algorithm}.

Then we prove that if  there exists a pure level set for an initialized model, our proposed approach can purify incorrect  labels and the classifier $f$ will finally match the Bayes optimal classifier $h$ after sufficient rounds $R$ under the instance-dependence PLL setting .
\begin{theorem}\label{theorem2}
	For any flipping probability of each incorrect label $\xi^{j}(\bm{x})$, define $e_0=\frac{(1+t)\alpha+\frac{\epsilon}{6}}{1+\alpha}$. And for a given function $f_0$ there exists a level set $L(e_0)$ which is pure for $f_0$. If one runs purification  in Theorem \ref{theorem1} with enough traing data ($ n \rightarrow \infty $) starting with $f_0$ and the initialization: (1) $e_0 \geq \frac{(1+t)\alpha+\frac{\epsilon}{6}}{1+\alpha}$, (2) $R \geq\frac{6l}{\epsilon}\log(\frac{1-\epsilon}{\frac1c-e_0})$, (3) $e_\text{end} \geq \epsilon$, then we have: 
	$$\mathbb{P}_{\bm{x}\sim D}[y_{f_{final}(\bm{x})}=h^\star(\bm x)]\geq1-c^{\star}\epsilon$$
\end{theorem}
The proof of Theorem \ref{theorem2} is provided in Appendix A.3. According to Theorem \ref{theorem2}, the learned classifier under the instance-dependent PLL setting will be consistent with the Bayes optimal classifier eventually. Theorem \ref{theorem2} shows that the classifier can be guaranteed to eventually approximate the Bayes optimal classifier.

\begin{table*}[t]
	\caption{Classification accuracy (mean$\pm$std) of each comparing approach on  benchmark datasets corrupted by the ID generation process. }
	\centering
	\label{performance_benchmark}
	\scalebox{1}{
		\begin{tabular}{cccccc}
			\toprule
			& MNIST & Kuzushiji-MNIST & Fashion-MNIST & CIFAR-10 & CIFAR-100 \\
			\midrule
			{\fid}   & \textbf{99.28$\pm$0.02\%} & \textbf{91.09$\pm$0.14\%} & \textbf{96.93$\pm$0.07\%} & \textbf{93.00$\pm$0.26\%} & \textbf{71.82$\pm$0.08\%}  \\ 
			\midrule
			{\vpll}    & 99.03$\pm$0.02\% & 90.15$\pm$0.02\% & 96.31$\pm$0.12\% & 92.01$\pm$0.09\% & 71.48$\pm$0.12\% \\
			{\rcr}    & 98.81$\pm$0.07\% & 90.62$\pm$0.22\% & 96.64$\pm$0.10\% & 86.11$\pm$0.43\% & 71.07$\pm$0.25\% \\
			{\pico}    & 98.76$\pm$0.04\% & 88.87$\pm$0.06\% & 94.83$\pm$0.17\% & 89.35$\pm$0.17\% & 66.30$\pm$0.24\% \\
			{\proden}  & 99.01$\pm$0.02\% & 90.48$\pm$0.14\% & 96.14$\pm$0.07\% & 78.87$\pm$0.26\% & 55.59$\pm$0.08\% \\ 
			{\rcpll}   & 99.09$\pm$0.09\% & 90.56$\pm$0.14\% & 96.17$\pm$0.08\% & 80.13$\pm$0.14\% & 56.41$\pm$0.17\% \\ 
			{\ccpll}   & 99.08$\pm$0.10\% & 90.40$\pm$0.20\% & 96.12$\pm$0.10\% & 76.17$\pm$0.11\% & 56.48$\pm$0.06\% \\ 
			{\lws}    & 98.98$\pm$0.05\% & 89.82$\pm$0.2\% & 93.23$\pm$0.08\% & 43.16$\pm$0.63\% & 49.63$\pm$0.12\% \\ 
			{\cavl}    & 98.95$\pm$0.05\% & 87.85$\pm$0.06\% & 95.84$\pm$0.06\% & 75.41$\pm$4.77\% & 58.17$\pm$0.11\% \\ 
			{\clpl} & 98.83$\pm$0.05\% & 90.21$\pm$0.08\% & 93.18$\pm$0.08\% & 51.61$\pm$0.39\% & 30.84$\pm$0.40\% \\ 
			\bottomrule
		\end{tabular}
	}
\end{table*}

\begin{table*}[t]
	\caption{Classification accuracy (mean$\pm$std) of each comparing approach on  the real-world datasets.  }
	\label{performance_real}
	\centering
	\resizebox{\linewidth}{!} {
		\begin{tabular}{cccccccc}
			\toprule
			&  Lost    &  BirdSong   &   MSRCv2   &  Mirflickr   &   Malagasy    & Soccer Player     &     Yahoo!News \\
			\midrule
			{\fid}     & \textbf{78.57$\pm$0.45\% } & \textbf{74.47$\pm$0.36\%} & 45.86$\pm$0.28\% & \textbf{61.09$\pm$0.10\%} & \textbf{72.29$\pm$0.33\%}  & 54.48$\pm$0.10\% & \textbf{66.38$\pm$0.07\%} \\
			\midrule				
			{\vpll}     & 76.87$\pm$0.86\%  &73.39$\pm$0.26\% & \textbf{49.97$\pm$0.43\%}& 59
			13$\pm$0.12\% & 69.44$\pm$0.06\% &  55.81$\pm$0.10\% & 66.26$\pm$0.13\% \\
			{\proden}   & 76.47$\pm$0.25\% & 73.44$\pm$0.12\% & 45.10$\pm$0.16\%  & 59.59$\pm$0.52\% & 69.34$\pm$0.09\%  &  54.05$\pm$0.15\% & 66.14$\pm$0.10\% \\
			{\rcpll}    & 76.26$\pm$0.46\%  & 69.33$\pm$0.32\% & 49.47$\pm$0.43\% & 58.93$\pm$0.10\% & 70.69$\pm$0.14\%  &  \textbf{56.02$\pm$0.59\%} & 63.51$\pm$0.20\% \\
			{\ccpll}    & 63.54$\pm$0.25\%  & 69.90$\pm$0.58\%& 41.50$\pm$0.44\% & 58.81$\pm$0.54\% & 69.53$\pm$0.34\%  &  49.07$\pm$0.36\% & 54.86$\pm$0.48\% \\
			{\lws}     & 73.13$\pm$0.32\%  & 51.45$\pm$0.26\% & 49.85$\pm$0.49\%& 54.50$\pm$0.81\% &  59.34$\pm$0.25\% &  50.24$\pm$0.45\% & 48.21$\pm$0.29\% \\
			{\cavl}     & 73.96$\pm$0.51\%  & 69.63$\pm$0.93\% & 46.62$\pm$1.29\%& 57.13$\pm$0.10\% & 65.82$\pm$0.06\%  &  52.92$\pm$0.40\% & 60.97$\pm$0.13\% \\
			{\clpl}     & 63.39$\pm$0.12\%  & 62.90$\pm$3.33\% & 37.8$\pm$0.71\%& 58.87$\pm$0.10\% & 64.25$\pm$0.29\%  &  48.23$\pm$0.03\% & 49.42$\pm$0.13\% \\
			\bottomrule
		\end{tabular}
	}
\end{table*}

\section{Experiments}
\subsection{Datasets}\label{sec:dataset}

We adopt five widely used benchmark datasets including \texttt{MNIST} \cite{lecun1998gradient}, \texttt{Kuzushiji-MNIST} \cite{clanuwat2018deep}, \texttt{Fashion-MNIST} \cite{xiao2017fashion}, \texttt{CIFAR-10} \cite{krizhevsky2009learning}, \texttt{CIFAR-100 }\cite{krizhevsky2009learning}. 
These datasets are manually corrupted into ID partially labeled versions. 
Specifically, we set the flipping probability of each incorrect label corresponding to an instance $\bm{x}$ by using the confidence prediction of a neural network trained using supervised data parameterized by $\hat{\bm{\theta}}$ \cite{xu2021instance}.
The flipping probability $\xi^j(\bm{x})=\frac{f_j(\bm{x};\hat{\bm{\theta}})}{\max_{{j}\in\bar{Y}}f_j(\bm{x};\hat{\bm{\theta}})}$, where $\bar{Y}_i$ is the set of all incorrect labels except for the true label of $\bm{x}_i$.  
The average number of candidate labels (avg. \#CLs) for each benchmark dataset corrupted by the ID generation process is recorded in Appendix A.4.

In addition, seven real-world PLL datasets which are collected from different application domains are used, including \texttt{Lost} \cite{Cour_2011}, \texttt{Soccer Player }\cite{Zeng_2013}, \texttt{Yahoo!News} \cite{Guillaumin_2010} from automatic face naming,\texttt{ MSRCv2} \cite{liu2012conditional} from object classification, \texttt{Malagasy} \cite{garrette2013learning} from POS
tagging, \texttt{Mirflickr} \cite{huiskes2008mir} from web image classification, and \texttt{BirdSong} \cite{Briggs_2013} from bird song classification. 
The average number of candidate labels (avg. \#CLs) for each real-world PLL dataset is also recorded in Appendix A.4.

\begin{table*}
	\caption{Classification accuracy (mean$\pm$std) of each comparing approach on  benchmark datasets corrupted by the ID generation process. }
	\centering
	\label{enhance_benchmark}
	\scalebox{1}{
		\begin{tabular}{cccccc}
			\toprule
			& MNIST & Kuzushiji-MNIST & Fashion-MNIST & CIFAR-10 & CIFAR-100 \\
			\midrule
			{\proden}  & 97.70$\pm$0.03\% & 87.60$\pm$0.23\% & 87.21$\pm$0.11\% & 76.77$\pm$0.63\% & 55.12$\pm$0.12\% \\ 
			{\proden+\fid}   & 97.87$\pm$0.04\% & 88.70$\pm$0.02\% & 87.62$\pm$0.04\% & 79.00$\pm$0.28\% & 57.68$\pm$0.14\% \\ 
			\midrule		
			{\rcpll}   & 97.72$\pm$0.02\% & 87.25$\pm$0.06\% & 87.06$\pm$0.14\% & 76.49$\pm$0.52\% & 55.18$\pm$0.70\% \\ 
			{\rcpll+\fid}   & 98.08$\pm$0.03\% & 87.78$\pm$0.09\% & 87.45$\pm$0.05\% & 78.89$\pm$0.17\% & 57.66$\pm$0.11\% \\ 
			\midrule
			{\ccpll}   & 97.25$\pm$0.11\% & 83.31$\pm$0.07\% & 86.01$\pm$0.13\% & 72.87$\pm$0.82\% & 55.56$\pm$0.23\% \\ 
			{\ccpll+\fid}   & 97.99$\pm$0.06\% & 83.98$\pm$0.10\% & 86.32$\pm$0.06\% & 77.03$\pm$0.58\% & 56.18$\pm$0.06\% \\ 
			\midrule
			{\lws}    & 96.80$\pm$0.07\% & 84.46$\pm$0.22\% & 86.25$\pm$0.01\% & 46.77$\pm$0.66\% & 48.00$\pm$0.16\% \\ 
			{\lws+\fid}   & 97.47$\pm$0.06\% & 84.71$\pm$0.07\% & 86.40$\pm$0.05\% & 48.54$\pm$0.04\% & 49.61$\pm$0.27\% \\ 
			\midrule
			{\cavl}     & 96.25$\pm$0.40\% & 79.38$\pm$0.69\% & 84.66$\pm$0.05\% & 62.69$\pm$1.65\% & 47.35$\pm$0.16\% \\
			{\cavl+\fid}   & 96.71$\pm$0.11\% & 79.83$\pm$0.12\% & 85.04$\pm$0.10\% & 63.12$\pm$0.23\% & 47.61$\pm$0.06\%   \\
			\midrule
			{\clpl}     & 96.11$\pm$0.21\% & 83.31$\pm$0.24\% & 83.16$\pm$0.25\% & 53.61$\pm$0.31\% & 22.31$\pm$0.11\% \\
			{\clpl+\fid}   & 96.51$\pm$0.22\% & 83.63$\pm$0.11\% & 83.71$\pm$0.15\% & 54.22$\pm$0.51\% & 23.37$\pm$0.29\%   \\
			\bottomrule
		\end{tabular}
	}
\end{table*}

\begin{table*}[t]
	\caption{Classification accuracy (mean$\pm$std) of each comparing approach on  the real-world datasets.  }
	\label{enhance_real}
	\centering
	\scalebox{0.85}{
		\begin{tabular}{cccccccc}
			\toprule
			&  Lost    &  BirdSong   &   MSRCv2   &  Mirflickr   &   Malagasy       & Soccer Player  &     Yahoo!News \\
			\midrule
			{\proden}   & 76.47$\pm$0.25\% & 73.44$\pm$0.12\% & 45.10$\pm$0.16\%  & 59.59$\pm$0.52\% & 69.34$\pm$0.09\%  &  54.05$\pm$0.15\% & 66.14$\pm$0.10\% \\
			{\proden+\fid}     & 78.57$\pm$0.45\%  & 74.47$\pm$0.36\% & 45.86$\pm$0.28\%& 61.09$\pm$0.10\% & 72.29$\pm$0.33\%  &  54.48$\pm$0.10\% & 66.38$\pm$0.07\% \\
			\midrule
			{\rcpll}    & 76.26$\pm$0.46\%  & 69.33$\pm$0.32\% & 49.47$\pm$0.43\% & 58.93$\pm$0.10\% & 70.69$\pm$0.14\%  &  56.02$\pm$0.59\% & 63.51$\pm$0.20\% \\
			{\rcpll+\fid}    & 78.56$\pm$0.45\%  & 70.77$\pm$0.26\% & 51.18$\pm$0.59\% & 59.65$\pm$0.52\% & 71.04$\pm$0.10\%  &  56.49$\pm$0.03\% & 63.86$\pm$0.22\% \\
			\midrule
			{\ccpll}    & 63.54$\pm$0.25\%  & 69.90$\pm$0.58\%& 41.50$\pm$0.44\% & 58.81$\pm$0.54\% & 69.53$\pm$0.34\%  &  49.07$\pm$0.36\% & 54.86$\pm$0.48\% \\
			{\ccpll+\fid}    & 65.47$\pm$0.93\% & 71.50$\pm$0.06\% & 43.21$\pm$0.43\% & 59.89$\pm$0.48\% & 71.19$\pm$0.40\%  &  49.36$\pm$0.02\% & 55.22$\pm$0.05\% \\
			\midrule
			{\lws}     & 73.13$\pm$0.32\%  & 51.45$\pm$0.26\% & 49.85$\pm$0.49\%& 54.50$\pm$0.81\% &  59.34$\pm$0.25\% &  50.24$\pm$0.45\% & 48.21$\pm$0.29\% \\
			{\lws+\fid}     & 75.30$\pm$0.26\%  & 52.35$\pm$0.26\% & 52.42$\pm$0.86\%& 55.46$\pm$0.27\% & 60.85$\pm$0.57  &  50.94$\pm$0.47\% & 48.6$\pm$0.12\% \\
			\midrule
			{\cavl}     & 73.96$\pm$0.51\%  & 69.63$\pm$0.93\% & 46.62$\pm$1.29\%& 57.13$\pm$0.10\% & 65.82$\pm$0.06\%  &  52.92$\pm$0.40\% & 60.97$\pm$0.13\% \\
			{\cavl+\fid}   & 75.32$\pm$0.11\%  & 70.13$\pm$0.22\% & 46.92$\pm$0.13\%& 58.63$\pm$0.48\% & 67.70$\pm$0.19\%  &  53.44$\pm$0.10\% & 61.37$\pm$0.11\%   \\
			\midrule
			{\clpl}     & 63.39$\pm$0.12\%  & 62.90$\pm$3.33\% & 37.8$\pm$0.71\%& 58.87$\pm$0.10\% & 64.25$\pm$0.29\%  &  48.23$\pm$0.03\% & 49.42$\pm$0.13\% \\
			{\clpl+\fid}  & 64.73$\pm$0.14\% & 64.06$\pm$0.48\% & 39.32$\pm$0.24\% & 60.31$\pm$0.27\% & 66.04$\pm$0.25\% & 49.11$\pm$0.21\% & 50.33$\pm$0.18\%  \\
			\bottomrule
		\end{tabular}
	}
\end{table*}

\begin{table*}[t]
	\caption{Classification accuracy (mean$\pm$std) of the ablation studies on the warm-up round.  }
	\label{ablation}
	\centering
	\begin{tabular}{ccccccc}
		\toprule
		Warm-up   rounds & 0            & 1            & 5            & 10           & 15           & 20            \\
		\midrule
		Lost             & 78.28$\pm$0.25\% & 78.42$\pm$0.25\% & 78.42$\pm$0.68\% & 78.57$\pm$0.44\% & 78.72$\pm$0.25\% & 78.42$\pm$0.51\%   \\
		BirdSong         & 74.10$\pm$0.35\% & 73.14$\pm$0.31\% & 74.10$\pm$0.35\% & 74.47$\pm$0.40\% & 74.24$\pm$0.40\% & 74.20$\pm$0.23\% \\
		MSRCv2           & 45.01$\pm$0.29\% & 44.91$\pm$0.43\% & 45.67$\pm$0.16\% & 45.58$\pm$0.29\% & 45.77$\pm$0.16\% & 45.67$\pm$0.33\%     \\
		Soccer Player    & 54.32$\pm$0.08\% & 54.38$\pm$0.05\% & 54.42$\pm$0.02\% & 54.44$\pm$0.03\% & 54.43$\pm$0.03\% & 54.42$\pm$0.05\%    \\
		Yahoo!News       & 66.25$\pm$0.08\% & 66.33$\pm$0.04\% & 66.31$\pm$0.04\% & 66.42$\pm$0.13\% & 66.4$\pm$0.15\%  & 66.39$\pm$0.17\%   \\
		Kuzushiji-mnist  & 88.31$\pm$0.12\% & 88.3$\pm$0.15\%  & 88.74$\pm$0.49\% & 88.58$\pm$0.36\% & 88.73$\pm$0.24\% & 88.87$\pm$0.29\%   \\
		Fashion-mnist    & 87.22$\pm$0.11\% & 87.27$\pm$0.04\% & 87.47$\pm$0.17\% & 87.54$\pm$0.18\% & 87.61$\pm$0.07\% & 87.63$\pm$0.03\%    \\
		\bottomrule
	\end{tabular}
\end{table*}

\subsection{Baselines}
The performance of {\fid} is compared against five deep PLL approaches: 
\begin{itemize}[topsep=0ex,leftmargin=*,parsep=1pt,itemsep=1pt]	
	\item  {\proden} \cite{lv2020progressive}: A progressive identification approach which approximately minimizes a risk estimator and  identifies the true labels in a seamless manner; 
	\item {\rcpll} \cite{feng2020provably}: A risk-consistent approach which employs the loss correction strategy to establish the true risk by only using the partially labeled data;
	\item {\ccpll} \cite{feng2020provably}: A classifier-consistent approach which also uses the loss correction strategy to learn the classifier that approaches the optimal one;
	\item {\vpll} \cite{yao2020deep}: An ID PLL approach which recovers the latent label distribution via variational inference methods;
	\item {\lws} \cite{wen2021leveraged}: A risk-consistent approach which proposes a leveraged weighted loss to trade off the losses on candidate labels and non-candidate ones.
	\item {\cavl}  \cite{zhang2021exploiting}: A progressive identification approach which exploits the class activation value to identify the true label in candidate label sets.
	\item {\clpl} \cite{Cour_2011}: A avearging-based disambiguation approach based on a convex learning formulation.
	\item {\pico} \cite{wang2022pico}: A data-augmentation-based method which identifies the true label via contrastive-learning with learned prototypes for image datasets.
	\item {\rcr} \cite{wu2022revisiting}: A data-augmentation-based method which identifies the true label via  consistency regularization with  random augmented instances for image datasets.
\end{itemize}

For the benchmark datasets, we use the same data augmentation strategy for the data-augmentation-free methods ({\vpll}, {\proden}, {\rcpll}, {\ccpll}, {\lws} and {\cavl}) to make fair comparisons with the data-augmentation-based methods ({\pico} and {\rcr}). However, data augmentation cannot be employed on the realworld datasets that contain extracted feature from audio and video data, we just compared our methods with the data-augmentation-free methods on realworld datasets.

For all the deep approaches, We used the same training/validation setting, models, and optimizer for fair comparisons. 
Specifically, a 5-layer LeNet is trained on {MNIST}, {Kuzushiji-MNIST} and {Fashion-MNIST}, the  Wide-ResNet-28-2 \cite{zagoruyko2016wide,YangZFJZ17} is trained on {CIFAR-10} and {CIFAR-100}, and the linear model is trained on real-world PLL datasets, respectively. 
The hyper-parameters are selected so as to maximize the accuracy on a validation set (10\% of the training set). We run 5 trials on the benchmark datasets and the real-world PLL datasets. The mean accuracy as well as standard deviation are recorded for all comparing approaches.  
All the comparing methods are implemented with PyTorch.

\subsection{Experimental Results}
Table~\ref{performance_benchmark} and Table~\ref{performance_real} report the classification accuracy of each approach on benchmark datasets corrupted by the ID generation process and the real-world PLL datasets, respectively. The best results are highlighted in bold.
Due to the inability of data augmentation to be employed on extracted feature , we didn't compare our methods with {\pico} and {\rcr} on real-world datasets.
As shown in Table~\ref{performance_benchmark} and Table~\ref{performance_real}, it is impressive to observe that:
\begin{itemize}[topsep=0ex,leftmargin=*,parsep=1pt,itemsep=1pt]	
	\item   {\fid} achieves the best performance against other approaches in most cases; 
	\item The performance advantage of {\fid} over comparing approaches is stable under varying the number of candidate labels.
	\item {\fid} achieves the best performance against other approaches on all benchmark datasets by the instance-dependence generation process.
	\item {\fid} achieves the best performance against other approaches on all real-world datasets except {\vpll} on \texttt{MSRCv2} and {\rcpll} on \texttt{Soccer Player}.
\end{itemize}

\begin{figure}[t]
	\centering
	\includegraphics[width=0.43\textwidth]{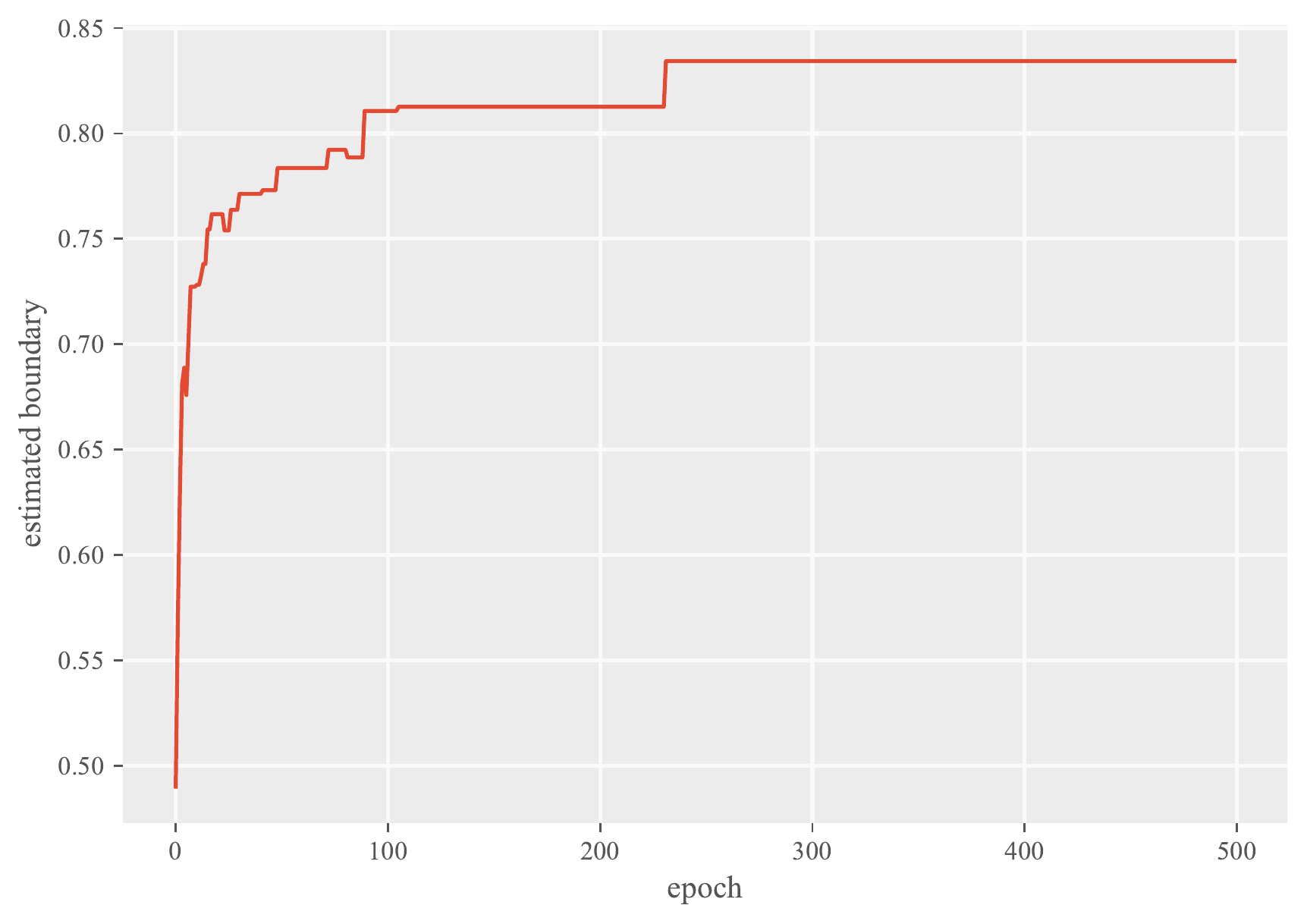}
	\caption{Estimated purified region on \texttt{Lost}.}\label{region}
\end{figure}

\begin{figure}[t]
	\centering
	\includegraphics[width=0.43\textwidth]{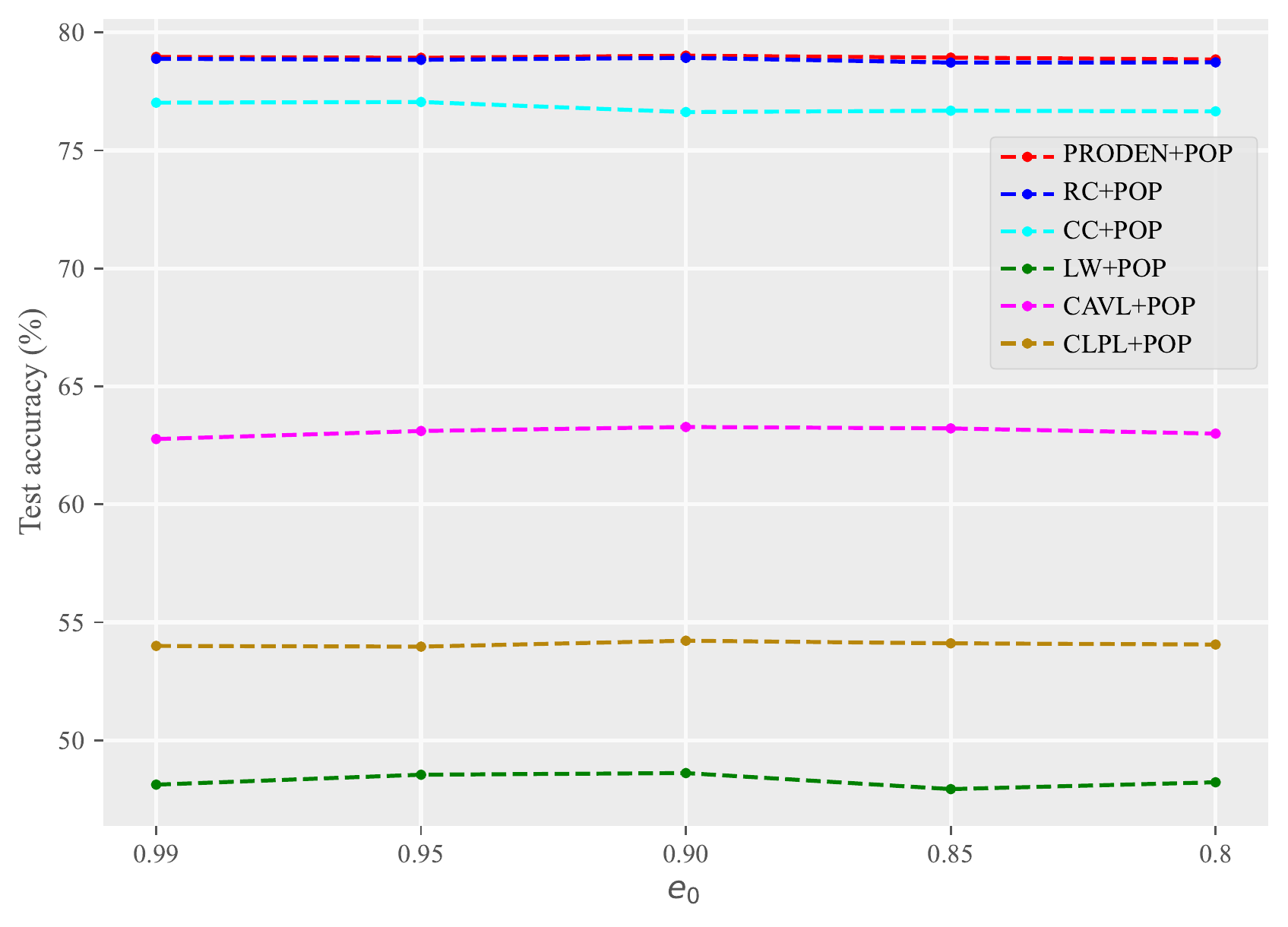}
	\caption{Hyper-parameter sensitivity on \texttt{CIFAR-10}.}\label{sensitivity}
\end{figure}

\subsection{Further Analysis}

In addition, to analysis the purified region in Theorem~\ref{theorem1}, we employ the confidence predictions of $f(\bm{x},\tilde{\bm{\theta}})$ (the network in Section~\ref{sec:dataset}) as the posterior and plot the curve of the estimated purified region  in every epoch on \texttt{Lost} in Figure \ref{region}. We can see that although the estimated purified region  would be not accurate  enough, the curve could show that  the trend of continuous increase for the purified region. 

As the framework of {\fid} is flexible for the loss function,  we integrate the proposed method with the previous methods for instance-independent PLL including {\proden}, {\rcpll}, {\ccpll}, {\lws}, {\cavl} and {\clpl}. In this subsection,  we empirically prove that the previous methods for instance-independent PLL could be promoted to achieve better performance  after integrating with {\fid}.

Table \ref{enhance_benchmark} and Table \ref{enhance_real}   report the classification accuracy of each  method for instance-independent PLL  and its variant integrated with {\fid} on benchmark datasets corrupted by the ID generating procedure and the real-world datasets, respectively. We didn't use any data augmentation on benchmark datasets in this part of experiments.   As shown in Table \ref{enhance_benchmark} and Table \ref{enhance_real}, the  approaches integrated with {\fid} including {\proden+\fid}, 	{\rcpll+\fid}, 	{\ccpll+\fid} ,	{\lws+\fid}, {\cavl+\fid} and {\clpl+\fid}  achieve superior performance against original method, which clearly validates the usefulness of {\fid} framework for improving performance for ID PLL.

Figure  \ref{sensitivity}  illustrates  the variant integrated with {\fid}  performs under different
hyper-parameter configurations on \texttt{CIFAR-10}  while similar observations are also
made on other data sets.  The hyper-parameter sensitivity  on other datasets could be founded in Appendix A.4. As shown in Figure  \ref{sensitivity}, it is obvious that the performance
of  the variant integrated with {\fid}  is relatively stable across a broad range of each
hyper-parameter. This property is quite desirable as {\fid} framework could
achieve robust classification performance. 

The ablation studies on the warm-up round are shown in Table  \ref{ablation}. These results show that the performance of our method does not rely on the warm-up strategy.

\section{Conclusion}
In this paper, the problem of partial label learning is studied  where  a  novel  approach  {\fid} is  proposed.  we consider ID partial label learning and  propose a theoretically-guaranteed  approach,  which could train the classifier with  progressive purification of the candidate labels and  is theoretically
guaranteed to eventually approximates the Bayes optimal classifier for ID PLL.  Experiments on benchmark and real-world datasets validate the effectiveness of the proposed method. 
If PLL methods become very effective, the need for exactly annotated data would be significantly reduced. 

\section{Acknowledgments}
This research was supported by the National Key Research \& Development Plan of China (No. 2021ZD0114202), the National Science Foundation of China (62206050, 62125602, and 62076063), China Postdoctoral Science Foundation (2021M700023), Jiangsu Province Science Foundation for Youths (BK20210220), Young Elite Scientists Sponsorship Program of Jiangsu Association for Science and Technology (TJ-2022-078), and the Big Data Computing Center of Southeast University.
\nocite{langley00}

\bibliography{xning}
\bibliographystyle{icml2023}

\newpage
\appendix
\onecolumn

\section{Appendix}
\subsection{Proofs of Theorem 1} 
Assume that  there exists a set $ L(e) $ for all $\bm{x} \in L(e)$ which satisfies $y^{\bm{x}}=\arg\max_{j} f_j(\bm x)$ and $ p(y^{\bm{x}}|\bm{x})- p(o|\bm{x})\geq e$, we have
\begin{equation}\label{1}
	\mathbb{E}_{(\bm{z},S)\sim \tilde{p}(\bm{z},S_{\text{new}})}\left[\mathbf{1}_{\{ {j \neq h^\star(\bm z)} \}}\Big\vert p(y^{\bm{z}}|\bm{z})- p(o|\bm{z})\geq p(y^{\bm{x}}|\bm{x})- p(o|\bm{x}) {,~j \in S_{\bm z}} \right]=0
\end{equation}
Let  $e_\text{new}$ be the new boundary and $\frac{\epsilon}{6l\alpha}( p(y^{\bm{x}}|\bm{x})-e)\leq e-e_\text{new}\leq\frac{\epsilon}{3l\alpha}( p(y^{\bm{x}}|\bm{x})-e)$. As the probability density function $d(u)$ of the margin $u(\bm{x})= p(y^{\bm{x}}|\bm{x})- p(o|\bm{x}) $ is bounded by  $c_{\star} < d(u) < c^{\star}$, we have the following result for $\bm{x}$ that satisfies $ p(y^{\bm{x}}|\bm{x})- p(o|\bm{x}) \geq e_\text{new}$ \footnote{Details of Eq. (3) in the paper submission}
\begin{equation}\label{2}
	\small
	\begin{aligned}
		&\mathbb{E}_{(\bm{z}, S)\sim \tilde{p}(\bm{z}, S_{\text{new}})}\left[\mathbf{1}_{\{ j \neq h^\star(\bm z) \}}\Big\vert {j \in S_{\bm z},} p(y^{\bm{z}}|\bm{z})- p(o|\bm{z})\geq p(y^{\bm{x}}|\bm{x})- p(o|\bm{x}) \right]\\
		\leq & \mathbb{E}_{(\bm{z}, S)\sim \tilde{p}(\bm{z}, S_{\text{new}})}\left[\mathbf{1}_{\{ j \neq h^\star(\bm z) \}}\Big\vert p(y^{\bm{z}}|\bm{z})- p(o|\bm{z})\geq p(y^{\bm{x}}|\bm{x})- p(o|\bm{x}) \right] \\
		=&\mathbb{P}_{\bm z}\left[ j \neq h^\star(\bm z) \Big\vert p(y^{\bm{z}}|\bm{z})- p(o|\bm{z})\geq p(y^{\bm{x}}|\bm{x})- p(o|\bm{x}) \right]\\
		=&\frac{\mathbb{P}_{\bm z}\left[ j \neq h^\star(\bm z), p(y^{\bm z} \vert \bm{z})-p(o \vert \bm{z}) \geq p(y^{\bm{x}} \vert \bm{x})-p(o \vert \bm{x}) \right]}{\mathbb{P}_{\bm z}\left[ p(y^{\bm z} \vert \bm{z})-p(o \vert \bm{z}) \geq p(y^{\bm{x}} \vert \bm{x})-p(o \vert \bm{x}) \right]}\\
		\leq& \frac{\mathbb{P}_{\bm z}\left[j \neq h^\star(\bm z), p(y^{\bm{z}} \vert \bm{z}) - p(o \vert \bm{z}) \geq e\right]}{\mathbb{P}_{\bm z}\left[ p(y^{\bm z} \vert \bm{z})-p(o \vert \bm{z}) \geq p(y^{\bm{x}} \vert \bm{x})-p(o \vert \bm{x})\right]}
		+\frac{\mathbb{P}_{\bm z}\left[ j \neq h^\star(\bm z), e_{\text{new}} \leq p(y^{\bm z} \vert \bm z)-p(o \vert \bm z) < e\right]}{\mathbb{P}_{\bm z}\left[ p(y^{\bm z} \vert \bm{z})-p(o \vert \bm{z}) \geq p(y^{\bm{x}} \vert \bm{x})-p(o \vert \bm{x}) \right]}\\
		= &\frac{\mathbb{P}_{\bm z}\left[j \neq h^\star(\bm z), p(y^{\bm{z}} \vert \bm{z}) - p(o \vert \bm{z}) \geq e\right]}{\mathbb{P}_{\bm z}\left[ p(y^{\bm z} \vert \bm{z})-p(o \vert \bm{z}) \geq e\right]} \frac{\mathbb{P}_{\bm z}\left[ p(y^{\bm z} \vert \bm{z})-p(o \vert \bm{z}) \geq e\right]}{\mathbb{P}_{\bm z}\left[ p(y^{\bm z} \vert \bm{z})-p(o \vert \bm{z}) \geq p(y^{\bm{x}} \vert \bm{x})-p(o \vert \bm{x}) \right]} \\
		& + \frac{\mathbb{P}_{\bm z}\left[  j \neq h^\star(\bm z) , e_{\text{new}} \leq p(y^{\bm z} \vert \bm z)-p(o \vert \bm z) < e\right]}{\mathbb{P}_{\bm z}\left[ p(y^{\bm z} \vert \bm{z})-p(o \vert \bm{z}) \geq p(y^{\bm{x}} \vert \bm{x})-p(o \vert \bm{x}) \right]}\\
		= &\underbrace{\mathbb{E}_{(\bm{z}, S)\sim \tilde{p}(\bm{z}, S)}\left[\mathbf{1}_{\{h(\bm z) \neq y^{\bm{z}}\}}\Big\vert p(y^{\bm z} \vert \bm{z})-p(o \vert \bm{z}) \geq e \right]}_{=0\text{(According to Eq. (\ref{1}))}} \frac{\mathbb{P}_{\bm z}\left[ p(y^{\bm z} \vert \bm{z})-p(o \vert \bm{z}) \geq e\right]}{\mathbb{P}_{\bm z}\left[ p(y^{\bm z} \vert \bm{z})-p(o \vert \bm{z}) \geq p(y^{\bm{x}} \vert \bm{x})-p(o \vert \bm{x}) \right]}\\
		& + \frac{\mathbb{P}_{\bm z}\left[j \neq y^{\bm{z}}, e_{\text{new}} \leq p(y^{\bm z} \vert \bm z)-p(o \vert \bm z) < e\right]}{\mathbb{P}_{\bm z}\left[ p(y^{\bm z} \vert \bm{z})-p(o \vert \bm{z}) \geq p(y^{\bm{x}} \vert \bm{x})-p(o \vert \bm{x}) \right]}\\
		= &\frac{\mathbb{P}_{\bm z}\left[ e_{\text{new}} \leq p(y^{\bm z} \vert \bm z)-p(o \vert \bm z) < e \right]}{\mathbb{P}_{\bm z}\left[ p(y^{\bm z} \vert \bm{z})-p(o \vert \bm{z}) \geq p(y^{\bm{x}} \vert \bm{x})-p(o \vert \bm{x}) \right]}\\
		\leq &\frac{c^{\star} (e - e_{\text{new}})}{c_{\star}\left(p(y^{\bm{x}} \vert \bm{x}) - e\right)}.
	\end{aligned}
\end{equation}
Due to that $\frac{\epsilon}{6l \alpha}(p(y^{\bm{x}} \vert \bm{x})-e)\leq e-e_{\text{new}} \leq\frac{\epsilon}{3l \alpha}(p(y^{\bm{x}} \vert \bm{x})-e)$ holds, we can further relax Eq. (\ref{2}) as follows: 
\begin{equation}
	\begin{aligned}
		&\mathbb{E}_{(\bm{z}, S)\sim \tilde{p}(\bm{z}, S_{\text{new}})}\left[\mathbf{1}_{\{ j \neq h^\star(\bm z) \}}\Big\vert j \in S_{\bm z}, p(y^{\bm z} \vert \bm{z})-p(o \vert \bm{z}) \geq p(y^{\bm{x}} \vert \bm{x})-p(o \vert \bm{x}) \right]\\
		&\leq \frac{c^{*} (e - e_{\text{new}})}{c_{*}\left(p(y^{\bm{x}} \vert \bm{x}) - e\right)}\\
		&\leq \frac{c^{*}}{c_{*}\left(p(y^{\bm{x}} \vert \bm{x}) - e\right)}\frac{\epsilon}{3l \alpha}(p(y^{\bm{x}} \vert \bm{x})-e)\\
		&= \frac{\epsilon}{3\alpha}.
	\end{aligned}
\end{equation}

Then, we can find that the assumption that the gap between $f_j(\bm x)$ and $f^\star_j(\bm x)$ should be controlled by the risk at point $\bm{z}$ implies:
\begin{equation}\label{4}
	\begin{aligned}
		&\left\vert f_j(\bm x) - f^\star_j(\bm z) \right\vert \\
		\leq & \, \alpha  \mathbb{E}_{(\bm{z}, S)\sim \tilde{p}(\bm{z}, S_{\text{new}})}\left[\mathbbm{1}_{\{h(\bm z) \neq y^{\bm{z}}\}}\Big\vert p(y^{\bm z} \vert \bm{z})-p(o \vert \bm{z}) \geq p(y^{\bm{x}} \vert \bm{x})-p(o \vert \bm{x}) \right] + \frac\epsilon6\\
		\leq & \, \alpha \frac{\epsilon}{3\alpha} + \frac\epsilon6\\
		\leq & \, \frac \epsilon2.
	\end{aligned}
\end{equation}
Hence, for $\bm{x}$ s.t. $p(y^{\bm{x}} \vert \bm{x})-p(o \vert \bm{x}) \geq e_{\text{new}}$, according to Eq. (\ref{4}) we have 
\begin{equation}
	\begin{aligned}
		f_{y^{\bm x}}(\bm x) - f_{j \neq y^{\bm x}}(\bm x) &\geq (p(y = y^{\bm x} \vert \bm x) - \frac \epsilon2) - (p(y = j \vert \bm x) + \frac \epsilon2)\\
		&= p(y = y^{\bm x} \vert \bm x) - p(y = j \vert \bm x) - \epsilon \\
		&\geq p(y = y^{\bm{x}} \vert \bm{x})-p(o \vert \bm{x}) - \epsilon \\
		& \geq e_{\text{new}} - \epsilon \\
		& \geq 0,
	\end{aligned}
\end{equation}

which means that $_j(\bm{x})$ will be the same label as $h^{\star}$ and thus the level set $L(e_{\text{new}})$ is pure for $f$. Meanwhile, the choice of $e_{\text{new}}$ ensures that 
\begin{equation}
	\begin{aligned}
		p(y^{\bm x} \vert \bm x)-e_{\text{new}} &\geq p(y^{\bm x} \vert \bm x) - (e - \frac{\epsilon}{6l\alpha}(p(y^{\bm x} \vert \bm x) - e)) \\
		& = p(y^{\bm x} \vert \bm x) - e + \frac{\epsilon}{6l\alpha}(p(y^{\bm x} \vert \bm x) - e) \\
		& = (1 + \frac{\epsilon}{6l\alpha})(p(y^{\bm x} \vert \bm x) - e).
	\end{aligned}
\end{equation}
Here, the proof of Theorem 1 has been completed.\\

\subsection{ Details of Eq. (5)}

If $f_m(\bm x) - f_{j \neq m} \geq e + \epsilon$, according to Eq. (\ref{4}) we have: 
\begin{equation}
	\begin{aligned}
		p(y^{\bm x} \vert \bm x)
		&\geq p(y=m \vert \bm x) \\
		&= p(y = j \vert \bm x) + p(y=m \vert \bm x) - p(y = j \vert \bm x) \\
		&\geq p(y = j \vert \bm x) + p(y = m \vert \bm x) - p(y = j \vert \bm x) \\
		&\geq p(y = j \vert \bm x) + (f_m(\bm x) - \frac{\epsilon}{2}) - (f_j(\bm x) + \frac{\epsilon}{2})\\
		&= p(y = j \vert \bm x) + (f_m(\bm x) - f_j(\bm x)) - \epsilon\\
		&\geq p(y = j \vert \bm x) + (e + \epsilon) - \epsilon\\
		& = p(y = j \vert \bm x) + e. \\
	\end{aligned}
\end{equation}

\begin{figure}[t]
	\centering
	\centering{\includegraphics[width=0.5\textwidth]{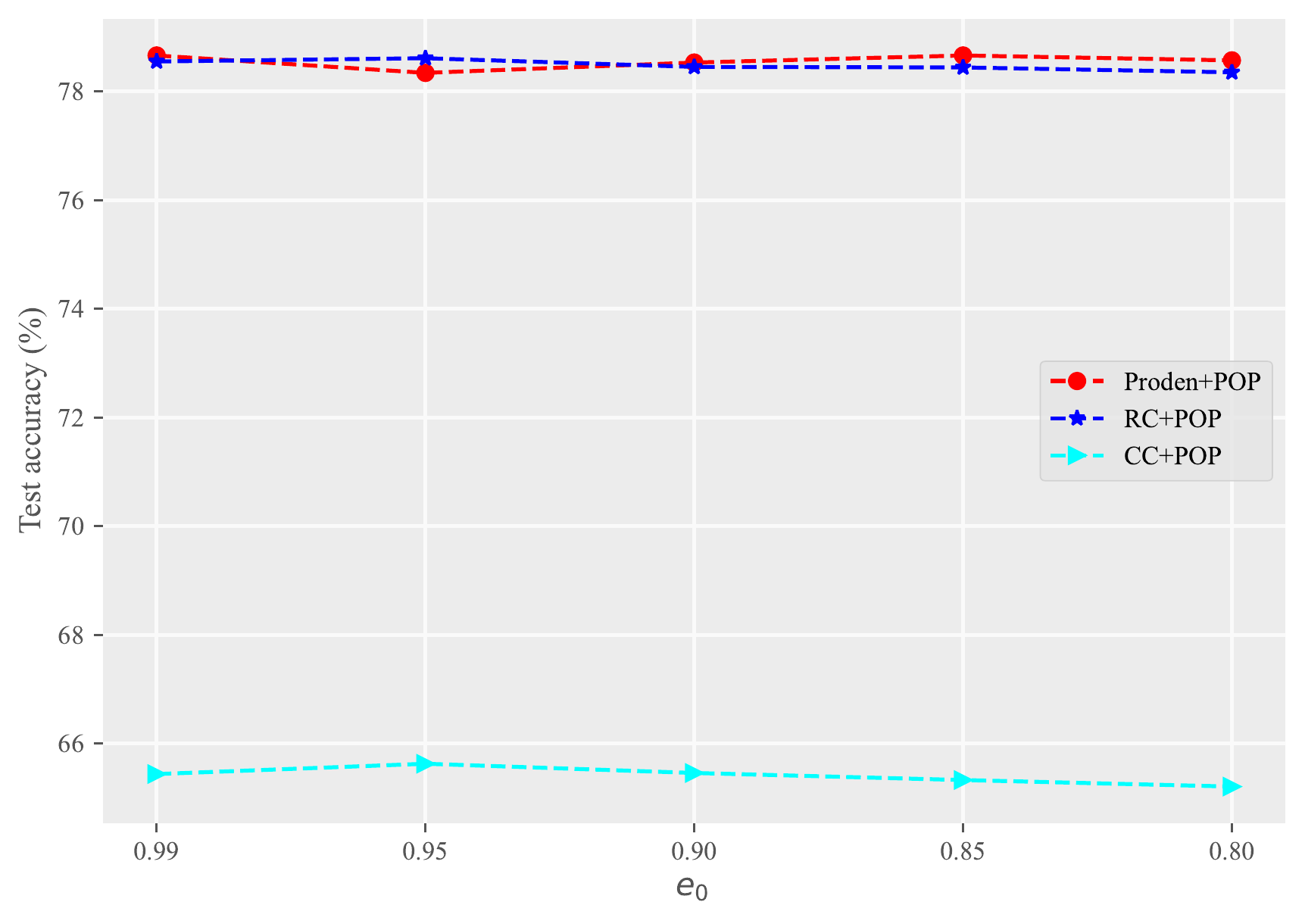}}
	\caption{Hyper-parameter sensitivity on {Lost}.}\label{sensitivity_2}
\end{figure}
\subsection{Proofs of Theorem 2}

To begin with, we prove that there exists at least a level set $L(e_0)$ pure to $f_0$. Considering $\bm{x}$ satisfies $p(y^{\bm x} \vert \bm{x})-p(o \vert \bm{x})\geq e_0$, we have $\mathbb{P}_{\bm{z}}\left[ j \neq h^\star(\bm z) \Big\vert j\in S_{\bm z}, p(y^{\bm z} \vert \bm{z})-p(o \vert \bm{z})\geq e_0 \right]\leq p(y^{\bm z} \vert \bm z) - e_0 + \xi^{j}(\bm z)$. Due to the assumption $
\vert f_j(\bm x) - f^\star_j(\bm x) \vert \leq \alpha  \mathbb{E}_{(\bm{z}, S)\sim \tilde{p}(\bm{z}, S)}\left[\mathbf{1}_{\{ j\neq h^\star(\bm z) \}}\Big\vert j\in S_{\bm z}, p(y^{\bm z} \vert \bm{z})-p(o \vert \bm{z}) \geq p(y^{\bm{x}} \vert \bm{x})-p(o \vert \bm{x}) \right] + \frac\epsilon6 $, it suffices to satisfy $\alpha(p(y^{\bm x} \vert \bm x)-e_0+\xi)+\frac \epsilon6 \leq e_0$ to ensure that $f_j(\bm{x})$ has the same prediction with $h^{\star}$ when $p(y^{\bm x} \vert \bm{x})-p(o \vert \bm{x}) \geq e_0$. Since we have $\xi^j(\bm{x})\leq p(y = j \vert \bm x)t \leq p(y^{\bm x} \vert \bm x)t$, by choosing $e_0\geq\frac{(1+t)\alpha  + \frac\epsilon6}{1+\alpha} \geq \frac{(1+t)\alpha p(y^{\bm x} \vert \bm x) + \frac\epsilon6}{1+\alpha}$ one can ensure that initial $f_{0}$ has a pure $L(e_0)$-level set. \\

Then in the rest of the iterations we ensure the level set $p(y^{\bm z} \vert \bm{z})-p(o \vert \bm{z})\geq e$ is pure. We decrease $e$ by a reasonable factor to avoid incurring too many corrupted labels while ensuring enough progress in label purification, i.e. $\frac{\epsilon}{6l \alpha}(p(y^{\bm{x}} \vert \bm{x})-e)\leq e-e_{\text{new}} \leq\frac{\epsilon}{3l \alpha}(p(y^{\bm{x}} \vert \bm{x})-e)$, such that in the level set $p(y^{\bm x} \vert \bm{x})-p(o \vert \bm{x})\geq e_{\text{new}}$ we have $\vert f_j(\bm{x})-f^\star_j(\bm x) \vert \leq \frac\epsilon2$. This condition ensures the correctness of flipping when $e \geq \epsilon$. The the purified region cannot be improved once $e < \epsilon$ since there is no guarantee that $f_j(\bm x)$ has consistent label with $h^{\star}$ when $p(y^{\bm x} \vert \bm{x})-p(o \vert \bm{x}) < \epsilon$ and $\vert f_j(\bm{x})-f^\star_j(\bm x) \vert \leq \frac\epsilon2$. To get the largest purified region, we can set $e_{\text{end}} = \epsilon$. Since the probability density function $d(u)$ of the margin $u(\bm{x}) = p(y^{\bm x}\vert \bm{x}) - p(o\vert \bm{x})$ is bounded by $c_{\star} \leq d(u) \leq c^{\star}$, we have: 
\begin{equation}
	\begin{aligned}
		\mathbb{P}_{\bm{x}\sim D}[y_{f_{final}(\bm{x})} \neq h^\star] &\leq \mathbb{P}[p(y^{\bm x} \vert \bm x) - p(o\vert \bm x) < e_{\text{end}}] \\
		& = \mathbb{P}_{\bm{x}\sim D}[p(y^{\bm x} \vert \bm x) - p(o\vert \bm x) < \epsilon] \\
		& \leq c^\star \epsilon.
	\end{aligned}
\end{equation}
Then $\mathbb{P}_{\bm{x}\sim D}[y_{f_{final}(\bm{x})} = h^\star] = 1 - \mathbb{P}_{\bm{x}\sim D}[y_{f_{final}(\bm{x})} \neq h^\star] \geq 1 - c^\star \epsilon$.

The rest of the proof is the total round $R \geq \frac{6\alpha l}{\epsilon}\log(\frac{1-\epsilon}{\frac1c - e_0})$, which follows from the fact that each round of label flipping improves the the purified region by a factor of $(1+\frac{\epsilon}{6l\alpha})$: 
\begin{equation}
	\begin{aligned}
		&\left(1+\frac{\epsilon}{6l \alpha}\right)^{R}\left(p(y^{\bm x} \vert \bm x)-e_0 \right) \geq p(y^{\bm x} \vert \bm x) -\epsilon \\
		&\Rightarrow \left(1+\frac{\epsilon}{6l \alpha}\right)^{R} \geq \frac{p(y^{\bm x} \vert \bm x) - \epsilon}{p(y^{\bm x} \vert \bm x)-e_{0}} \\
		&\Rightarrow R\log \left(1+\frac{\epsilon}{6l \alpha}\right) \geq \log \left(\frac{p(y^{\bm x} \vert \bm x) - \epsilon}{p(y^{\bm x} \vert \bm x)-e_{0}}\right) \\
		&\Rightarrow R\frac{\epsilon}{6l \alpha} \geq R\log \left(1+\frac{\epsilon}{6l \alpha}\right) \geq \log \left(\frac{p(y^{\bm x} \vert \bm x) - \epsilon}{p(y^{\bm x} \vert \bm x)-e_{0}}\right) \\
		&\Rightarrow R \geq\frac{6l \alpha}{\epsilon} \log \left(\frac{p(y^{\bm x} \vert \bm x) - \epsilon}{p(y^{\bm x} \vert \bm x)-e_{0}}\right) \geq \frac{6l \alpha}{\epsilon} \log (\frac{1 - \epsilon}{\frac1c-e_{0}}). \\			
	\end{aligned}
\end{equation}

\begin{table}[t]
	\caption{Characteristic of the benchmark datasets corrupted by the ID generation process.}
	\label{benchmark}
	\centering
	\renewcommand\arraystretch{1.25}
	\setlength{\tabcolsep}{2mm}{
		\begin{tabular}{c|c|c|c|c|c}
			\toprule
			\textbf{Dataset} & \textbf{\#Train} & \textbf{\#Test} & \textbf{\#Features} & \textbf{\#Class Labels} & \textbf{avg. \#CLs} \\\hline
			MNIST & 60000  & 10000   & 784    & 10 & 8.71 \\\hline
			Fashion-MNIST & 60,000  & 10,000   & 784    & 10   & 3.46 \\\hline	
			Kuzushiji-MNIST & 60,000  & 10,000   & 784    & 10   & 3.87 \\\hline
			CIFAR-10 & 50,000  & 10,000   & 3,072    & 10  & 3.68  \\\hline
			CIFAR-100  & 50,000  & 10,000   & 3,072    & 100& 4.64 \\
			\bottomrule
	\end{tabular}}
\end{table}

\begin{table}[t]
	\caption{Characteristic of the real-world PLL datasets.}
	\scriptsize
	\label{real-world}
	\centering
	\renewcommand\arraystretch{1.25}
	\setlength{\tabcolsep}{1mm}{
		\begin{tabular}{c|c|c|c|c|c|c}
			\toprule
			\textbf{Dataset} & \textbf{\#Train} & \textbf{\#Test} & \textbf{\#Features} & \textbf{\#Class Labels} & \textbf{avg. \#CLs} & \textbf{Task Domain}\\\hline
			Lost  & 898 &224  & 108   & 16    & 2.23 & automatic face naming \cite{Cour_2011}\\\hline		
			MSRCv2 & 1,406 &352  & 48    & 23    & 3.16 & object classification \cite{liu2012conditional}\\\hline
			Mirflickr & 2224 & 556 & 1536 & 14 & 2.76 & web image classification \cite{huiskes2008mir} \\\hline
			BirdSong & 3,998 &1,000  & 38    & 13    & 2.18 & bird song classification \cite{Briggs_2013}\\\hline
			Malagasy & 4243 & 1069 & 384 & 44 & 8.35 & POS Tagging \cite{garrette2013learning} \\\hline
			Soccer Player & 13,978 &3,494 & 279   & 171   & 2.09 & automatic face naming \cite{Zeng_2013}\\\hline
			Yahoo! News & 18,393 &4,598 & 163   & 219   & 1.91 & automatic face naming \cite{Guillaumin_2010}\\
			\bottomrule
	\end{tabular}}
\end{table}

\subsection{Details of Experiments}
We collect five widely used benchmark datasets including MNIST \cite{lecun1998gradient}, Kuzushiji-MNIST \cite{clanuwat2018deep}, Fashion-MNIST \cite{xiao2017fashion}, CIFAR-10 \cite{krizhevsky2009learning}, CIFAR-100 \cite{krizhevsky2009learning}. In addition, seven real-world PLL datasets which are collected from different application domains are used, including Lost \cite{Cour_2011}, Soccer Player \cite{Zeng_2013}, Yahoo!News \cite{Guillaumin_2010} from automatic face naming,, MSRCv2 \cite{liu2012conditional} from object classification, Malagasy \cite{garrette2013learning} from POS
tagging, Mirflickr \cite{huiskes2008mir} from web image classification, and BirdSong \cite{Briggs_2013} from bird song classification.   Figure  \ref{sensitivity_2}  illustrates  the variant integrated with {\fid}  performs under different
hyper-parameter configurations on {Lost}. 

The average number of candidate labels (avg. \#CLs) for each benchmark dataset corrupted by the ID generation process is recorded in Table-\ref{benchmark} and the average number of candidate labels (avg. \#CLs) for each real-world PLL dataset is recorded in Table-\ref{real-world}.

\end{document}